\documentclass[letterpaper, 10 pt, conference]{ieeeconf}
\IEEEoverridecommandlockouts
\let\labelindent\relax
\usepackage{enumitem}
\usepackage{balance}
\usepackage[font={small}]{caption}
\usepackage{subcaption}
\usepackage{array}
\usepackage{textcomp}
\usepackage{mathtools, nccmath}
\usepackage{graphicx}
\usepackage{amsfonts}
\usepackage{amsmath}
\usepackage{amssymb}
\usepackage{algorithm}
\usepackage{algorithmic}
\usepackage{hyperref}
\usepackage{tikz}
\usepackage{arydshln}
\usepackage{multirow}
\usepackage{bm}
\usepackage{epstopdf}
\usepackage{cite}
\usepackage{siunitx}

\usepackage{todonotes}

\DeclareMathOperator{\diag}{diag}

\usepackage{amsthm}

\newtheoremstyle{mystyle}
  {}
  {}
  {}
  {}
  {\bfseries}
  {.}
  { }
  {\thmname{#1}\thmnumber{ #2}\thmnote{ (#3)}}

\theoremstyle{mystyle}

\title{\LARGE \bf
Bracing for Impact: Robust Humanoid Push Recovery and Locomotion with Reduced Order Models
\thanks{All authors are affiliated with the AMBER Lab at the Department of Mechanical and Civil Engineering, California Institute of Technology, Pasadena, USA, \{lzyang, bwerner, aghansah, ames\}@caltech.edu.
Funding provided by the Technology Innovation Institute.} 
}

\author{Lizhi Yang, Blake Werner, Adrian B.Ghansah, Aaron D. Ames
}

\begin{document}

\maketitle
\thispagestyle{empty}
\pagestyle{empty}
\begin{abstract}
Push recovery during locomotion will facilitate the deployment of humanoid robots in human-centered environments. 
In this paper, we present a unified framework for walking control and push recovery for humanoid robots, leveraging the arms for push recovery while dynamically walking.  
The key innovation is to use the environment, such as walls, to facilitate push recovery by combining Single Rigid Body model predictive control (SRB-MPC) with Hybrid Linear Inverted Pendulum (HLIP) dynamics to enable robust locomotion, push detection, and recovery by utilizing the robot's arms to brace against such walls and dynamically adjusting the desired contact forces and stepping patterns. 
Extensive simulation results on a humanoid robot demonstrate improved perturbation rejection and tracking performance compared to HLIP alone, with the robot able to recover from pushes up to 100N for 0.2s while walking at commanded speeds up to 0.5m/s. 
Robustness is further validated in scenarios with angled walls and multi-directional pushes. 
\end{abstract}

\section{Introduction}\label{sec:introduction}

 It has been long understood that humans perform motor tasks in a robust and efficient way \cite{farley1996leg}, therefore, dynamic human behaviors can serve as inspiration for the control and design of bipedal robots \cite{ames2014human}.
 Humanoid robots have the potential of being agile and robust in complex environments \cite{atkeson2015no}, but require carefully designed sensing, decision and control frameworks.
The ultimate goal is to enable them to perform human-like motions and tasks, and be applied in multiple scenarios such as search and rescue, disaster response, and entertainment \cite{saeedvand2019comprehensive}.
However, model-based controllers capable of robust perturbation rejection have yet to be fully developed---especially those that leverage the arms while locomoting for dynamic push recovery. 

Push recovery is a fundamental requirement for a humanoid robot to maintain balance and stability, or to minimize damage, in the presence of external perturbations.
There are two main approaches: the first is to build a robust locomotion controller to reject external disturbance up to a certain level, as is done in many locomotion controllers \cite{alcaraz2013robust,li2020animated,li2021force, xiong2021robust,xiong20223,li2024reinforcement, khazoom2024tailoring, li2025gait}; the second is to design a specific control mode to recover from a push \cite{ogata2007falling,wang2018unified, stephens2010push, luo2015learning,ferigo2021emergence, choe2023seamless, marcucci2017approximate, lin2020robust, khazoom2022humanoid}. 
This is the focus of the current work.
In particular, we explore the use of a humanoid robot's arms to help itself recover from a push by providing additional contacts in the environment. 
We consider walls in the environment as they are common features that can provide stable support for a humanoid robot to brace itself against during push recovery. 
By leveraging walls, a robot can expand its recovery strategies and improve its chances of maintaining balance under perturbations.

We formulate the push recovery problem as a  model predictive control (MPC) problem, where the robot is allowed to use its arms to brace against walls in the environment.
As the full-order dynamics of the robot is high-dimensional and complex, we employ two reduced order models (ROMs) to simplify the problem: the Single Rigid Body (SRB) model \cite{di2018dynamic} is used to construct a generalizable model of the robot extendable with many possible contact forces at different contact sites, and the Hybrid Linear Inverted Pendulum (HLIP) model is used in order to robustly generate support polygons to provide additional stability assistance to the SRB model. 
By combining these two, we propose a unified framework to perform push detection, push recovery, and locomotion control.
We further demonstrate the effectiveness of the proposed method and categorize its performance in a high-fidelity simulation environment \cite{todorov2012mujoco}.
\begin{figure}
    \centering
    \includegraphics[width=\linewidth]{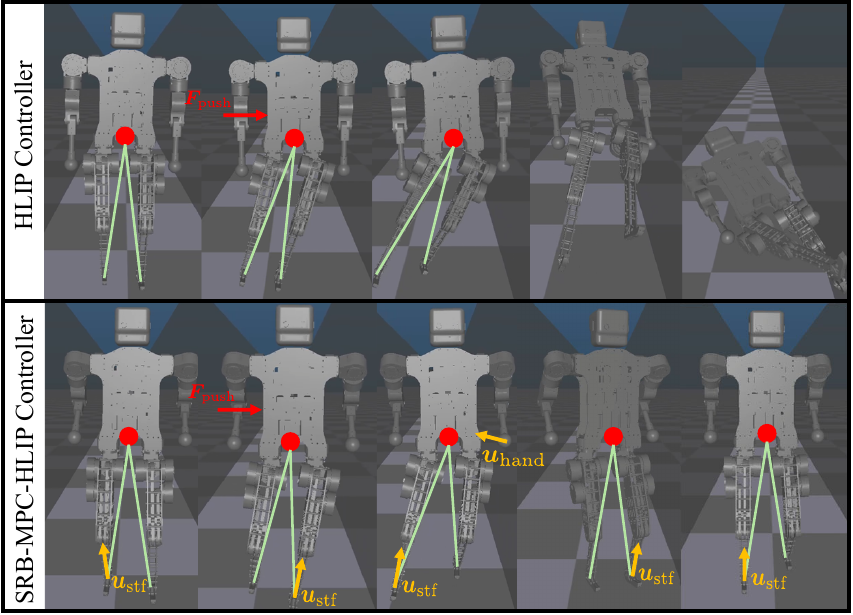}
    \caption{The proposed SRB-MPC-HLIP controller successfully utilizes the arms of the humanoid robot and adjusts its support polygon dynamically to brace the robot against walls in the event of a sudden push, while the HLIP controller could not account for such a sudden perturbation and fails to regain stable locomotion.}
    \label{fig:head}
\end{figure}
\begin{figure*}[htbp]
    \centering
    \includegraphics[width=\textwidth]{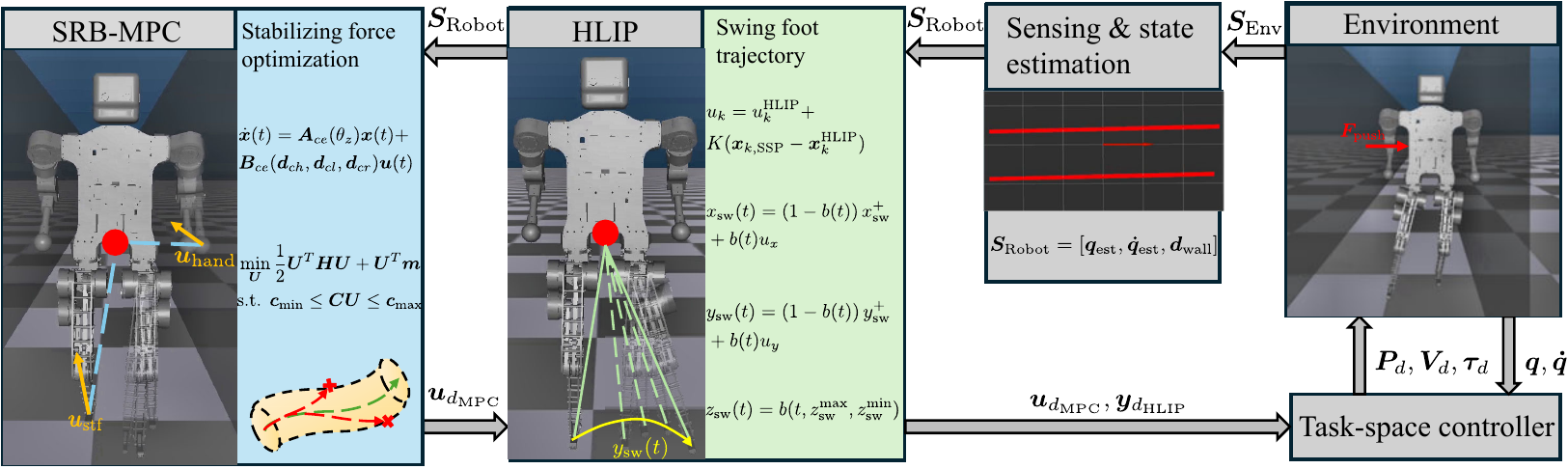}
    \caption{Framework for the SRB-MPC-HLIP controller. The robot runs state estimation and wall detection based on the sensor data from the environment as illustrated by the red lines in the sensing \& state estimation module, and the SRB-MPC and HLIP modules runs in tandem to reject large perturbations utilizing the environment by activating the recovery mode when necessary.}
    \label{fig:archi}
\end{figure*}
\subsection{Related Work}
\subsubsection{Bipedal locomotion}
Bipedal locomotion has long been a topic of interest in the robotics field, and has been studied extensively in the past few decades \cite{reher2021dynamic}.
Multiple strategies exist to reject certain amounts of perturbation.
One straight-forward approach is to adjust the angular momentum of the robot to counter the disturbance \cite{alcaraz2013robust}, however, this method depends heavily on the high-torque output of the joint motors.
In \cite{li2020animated}, a robust 3D locomotion controller is constructed using Hybrid Zero Dynamics (HZD) gait libraries tracked online with PD control, but its robustness relies on the size and granularity of the gait library.
\cite{li2021force} uses SRB-MPC and \cite{xiong2021robust,xiong20223} use HLIP to construct robust locomotion controllers respectively, but are constrained by the model inaccuracy caused by deviation from their nominal trajectories.
 A reinforcement-learning-based framework is proposed in \cite{li2024reinforcement}, but requires a large amount of training data, carefully designed training curriculum, and heavy domain randomization.
\subsubsection{Push recovery}\label{sssec:push_recovery}
Push recovery is currently an active area of research for humanoid robotics.
Depending on the effect of the push, it can be categorized into two types: mitigating the damage caused by the fall subsequent to the push \cite{ogata2007falling,wang2018unified}, or providing more robustness to reject the push and continue its original task \cite{stephens2010push, luo2015learning,ferigo2021emergence, choe2023seamless, marcucci2017approximate, lin2020robust, khazoom2022humanoid, ding2022orientation, kim2023model, dallard2024robust} (the focus of this paper).
The authors of \cite{stephens2010push} construct a stepping controller to reject pushes by bringing the robot center of mass (CoM) into a new support polygon.
\cite{luo2015learning,ferigo2021emergence} propose learning-based methods for whole-body control to recover from a push, at the expense of requiring large footprints for the former and unachievable torque spikes for the latter.
\cite{choe2023seamless} synthesizes a hip-ankle-step strategy to recover from a push, but makes the assumption of directly applying angular acceleration to the robot body for orientation adjustment.
In \cite{marcucci2017approximate}, a multi-contact method is proposed that utilizes the arms of the robot to interact with the environment to reject pushes, however, the method requires offline sampling for mode sequences, thus limiting its adaptability. 
\cite{lin2020robust} introduces a contact sequence planner using iterative kino-dynamic optimization and a neural network captureability classifier, but requires large computational resources and is difficult to implement real-time.
\cite{khazoom2022humanoid} proposes a hierarchical MPC method that uses the humanoid's arms to reject disturbance and \cite{ding2022orientation} leverages orientation dynamics, but both do not extend into locomotion scenarios when discussing push recovery or interact with the environment.
\cite{kim2023model} utilizes capture-point MPC and a stepping controller to maintain stability while experiencing stable pushes, but again, it does not extend to locomotion scenarios.
Similarly, \cite{dallard2024robust} also only discusses push recovery without the use of the humanoid's arms.

To address the aforementioned issues in the current literature, we attempt to unify push recovery and locomotion control by utilizing a combination of SRB-MPC for multi-contact recovery and HLIP for support polygon adjustment.

\subsection{Contributions}
The main contributions of this work are as follows: (1) we propose a novel SRB-MPC-HLIP controller to dynamically account for locomotion in the context of perturbation rejection by  utilizing the humanoid's arms to dynamically interact with the environment and adjusting the support polygon to reject moderate deviation of the CoM during active locomotion. (2) we construct a unified  framework capable of environment understanding, fall detection, reactive contact scheduling, and push recovery to demonstrate and quantify its effectiveness in a high-fidelity simulation.  
Namely, we subject a humanoid robot to pushes of up to 100N at different locations on the torso during different phases of locomotion, from which the robot is able to recover using variable wall geometries. 
We find that our approach enables robust disturbance rejection while dynamically locomoting through the use of the humanoid robot's arms, providing new evidence on the benefit of model-based methods in achieving human-like recovery behaviors.

\section{Background}\label{sec:background}
In this section, we provide a brief overview of concepts and methods that are relevant to our work. We first introduce the concept of push recovery, then discuss the use of the SRB and HLIP models to reduce problem complexity. We also discuss the use of SRB-MPC for control. 
\subsection{Push Recovery}
Push recovery is the ability of a robot to maintain balance and stability after experiencing an external perturbation. 
It is particularly important for humanoid robots because they are inherently unstable and are susceptible to disturbances. 
We define a push as a sudden force applied to the robot that causes it to deviate from its nominal stable trajectory, usually caused by an external agent, such as a human or other robots.
If the robot identifies its current state as non-recoverable when employing its nominal control policy, it must generate a new control policy on-the-fly, and execute the new policy to recover from the push, all within a short time frame, typically less than 0.5 seconds \cite{mungai2024fall}.
Due to the limited time frame in which a robot has to detect its state and generate a stabilizing policy, push recovery remains a challenging problem.
Current approaches are as discussed in \ref{sssec:push_recovery}.
Here, we consider the case of a humanoid robot that is capable of using its arms to brace against the environment to recover from a push, and continue locomotion.
\subsection{Single Rigid Body Model}\label{ssec:SRB}
In order to simplify the control problem and reduce the computational complexity to run the controller in real-time, we use the SRB model \cite{stephens2010push, di2018dynamic}.
The robot is modeled as a single rigid body that is manipulated by a set of forces and moments at the contact sites.
The SRB model is reasonable to use for the humanoid robot in Fig.\ref{fig:head} as its mass and inertia are concentrated around the torso, and the arms and legs are relatively light.
For each contact site, we define a contact wrench $\bm{u}_i = [\bm{F}_i^T, \bm{M}_i^T]^T$, where $\bm{F}_i$ is the contact force and $\bm{M}_i$ is the contact moment.
Thus, the rigid body dynamics can be written as
\begin{equation}
\resizebox{0.45\hsize}{!}{$
    \begin{aligned}
        \bm{\ddot{p}}_c &= \frac{1}{m}\sum_{i=1}^{n} \bm{F}_i - \bm{g}\\
        \frac{d}{dt}(\bm{I}_{G}{\bm{\omega}})&= \sum_{i=1}^{n} {(\bm{d}^{\times}_{ci}\bm{F}_i+\bm{M}_i)}\\ 
        \dot{\bm{R}} &= \bm{\omega}^{\times} \bm{R}
    \end{aligned}$}
\end{equation}
where $\bm{p}_c$ is the position of the center of mass, $\bm{I}_G$ is the inertia matrix, $\bm{\omega}$ is the angular velocity, $\bm{R}$ is the orientation matrix from the robot frame to the world frame, $\bm{g}$ is the gravity vector pointing towards earth, and $\bm{d}_{ci}=\bm{p}_c-\bm{p}_i$ is the vector from the robot's CoM to contact site $i$.
We also use ${(\cdot)}^{\times}$ to represent the skew operator. 
Due to the existence of non-linearities, we employ the linear approximations in \ref{sec:methods} to formulate the MPC problem.


\subsection{Hybrid Linear Inverted Pendulum Model}
As the SRB model only considers limbs in contact with the environment, we use the HLIP model\cite{xiong2021robust} to generate viable swing foot trajectories robust to perturbed CoM velocities, such as those caused by an external force. 
This robustness is a consequence of the variable support polygon constructed by the varying step sizes: subject to a perturbation force, the step size in the requisite direction increases and the polygon size in that direction does so as well. 
The result is robust and performant push recovery through stepping even before utilizing the SRB-MPC. 

The HLIP model describes periodic trajectories of the robot's CoM during locomotion by utilizing a ROM where the robot is modeled as a planar linear inverted pendulum with the following dynamics during the single support (SSP), or swing phase, as:

\begin{equation}
    \underbrace{\frac{d}{dt} \begin{bmatrix}
        p \\ \dot{p}
    \end{bmatrix}}_{\dot{x}_{\text{SSP}}} = \underbrace{\begin{bmatrix}
        0 & 1 \\
        \lambda^2 & 0
    \end{bmatrix}}_{A_{\text{SSP}}} \underbrace{\begin{bmatrix}
        p \\ \dot{p}
    \end{bmatrix}}_{x_{\text{SSP}}}
\end{equation}

and during the double support (DSP), or stance phase, as:  
\begin{equation}
    \ddot{p} = 0
\end{equation}

where $p$ is the CoM position for the HLIP model in the sagittal or frontal plane respectively, $\lambda = \frac{g}{z_0}$, and $z_0$ is the nominal height of the CoM\cite{xiong20223}.
Furthermore, the model defines the Step-to-Step (S2S) dynamics as the discrete states between each step and how they evolve. 
Using a combination of the simplified model and the discrete dynamics, we can then define a feedback law on the step lengths in order to achieve desired CoM positions and velocities.

\subsection{Model Predictive Control} \label{ssec:MPC-II}
We can formulate the control problem as an MPC framework composed of a sequential quadratic program over the discretized dynamics of the SRB model:
\begin{equation}
\resizebox{0.75\hsize}{!}{$
    \begin{aligned}
    \min_{\bm{u}} & \sum_{k=0}^{N-1} ||\bm{x}[k+1] - \bm{x}_{\text{ref}}[k+1]||_{\bm{Q}}
    + ||\bm{u}[k]||_{\bm{R}}\\
    \text{subject to} & \quad \bm{x}[k+1] = {\bm{A}}_{d}[k] \bm{x}[k] +  {\bm{B}}_{d}[k] \bm{u}[k] \\
    & \quad \bm{c}_{\text{min}} \leq \bm{C}[k]\bm{u}[k] \leq \bm{c}_{\text{max}} \\
    & \quad \bm{D}[k]\bm{u}[k] = 0 \\
    & \quad \bm{x}[0] = \bm{x}_{\text{init}}
    \end{aligned}$}
\end{equation}
where $\bm{x}[k]$ is the predicted state at the $k$-th time step, $\bm{x}_\text{ref}$ is the reference trajectory, $\bm{x}_\text{init}$ is the current state, $\bm{u}$ is the control input, $ {\bm{A}}_{d}$ and $ {\bm{B}}_{d}$ are the discretized state transition matrices, $\bm{C}$ and $\bm{D}$ are the inequality and equality constraint matrices for input and state constraints, and $\bm{Q}$ and $\bm{R}$ are the state and control input cost weight matrices.
The optimal control inputs are then applied to the robot through a task-space to joint-space mapping, and the optimization is repeated at each time step.

\section{Multi-contact Unified Controller for Locomotion and Push Recovery}\label{sec:methods}
In this section, we discuss the control architecture and the SRB-MPC-HLIP formulation used for push recovery.
We first provide an overview of the control architecture, then discuss the formulation of the controller in detail.
\subsection{Controller Architecture}\label{ssec:recovery}
During locomotion, the robot keeps track of its proprioceptive states, i.e. the generalized positions $\bm{q}$ and velocities $\dot{\bm{q}}$ by aggregating data from onboard odometry and joint sensors and passing it through a Kalman filter.
It also perceives the environment around it, namely the distance to the walls $\bm{d}_{\text{wall}}$, by running line detection algorithms using LiDAR scans based on \cite{pfister2003weighted}. 
The controller consists of two modes: normal locomotion mode and recovery mode.
During normal locomotion, the SRB and HLIP models work in tandem to generate the required stance foot ground reaction wrenches $\bm{u}_{d_{\text{MPC}}}$ and swing foot trajectories to track desired velocity commands.
If perturbed, we activate the recovery mode if (1) the projected CoM states from the SRB-MPC deviate significantly from the reference states, namely if the predicted $\bm{\dot{p}}$ and $\bm{\omega}$ diverge more than 0.4 m/s and 0.2 rad/s from the reference respectively based on emperical testing and (2) the workspace of the arms can reach the walls.
The recovery mode consists of both a hand strategy and a stepping strategy. For the hand strategy, an inverse kinematics controller moves the hand to a specified position on the wall relative to the robot frame. 
Once the robot's hand makes contact with the wall as determined by the state estimation module, contact force constraints are updated based on which hand is in contact, and the optimal feed-forward torque is mapped from the desired force from the SRB model.  
For the stepping strategy, the robot increases its stepping frequency from 3Hz to 5Hz when in the recovery state to quickly adjust its support polygon and CoM states.
As $\bm{d}_{\text{wall}}$ is known, the robot is aware of possible collisions of its feet to the walls and limits its next step choice location accordingly.
The entire control architecture is shown in Fig. \ref{fig:archi}.

\subsection{Robot Dynamics}
\label{sub:dynamics}
We quantize the stability of the robot with the CoM states, namely, the CoM position $\bm{p}_c=[x_c,y_c,z_c]^T$, linear velocity $\dot{\bm{p}}_c=[\dot{x}_c,\dot{y}_c,\dot{z}_c]^T$, orientation $\bm{\theta}=[\theta_x,\theta_y,\theta_z]^T$, and angular velocity $\bm{\omega}=[\omega_x,\omega_y,\omega_z]^T$ in the world frame.
In order to control these states, we use the SRB model to simplify the full-order dynamics of the robot as discussed in \ref{ssec:SRB}.
Consider the inputs
\begin{equation}
\resizebox{0.65\hsize}{!}{$
\begin{aligned}
    \bm{u} &= \left[\begin{array}{cccccc} \bm{F}_{h} & \bm{F}_{l} & \bm{F}_{r} & \bm{M}_{h} & \bm{M}_{l} & \bm{M}_{r} \end{array}\right]^T \\
\end{aligned}$}
\end{equation}
where $\bm{F}_i$ and $\bm{M}_i$ are the contact forces and moments,
and subscripts $h$, $l$, and $r$ denote the hand, left foot, and right foot respectively.
Thus we have a linear mapping between the inputs and the CoM acceleration and angular momentum:
\begin{equation}
\resizebox{0.45\hsize}{!}{$    \begin{aligned}
        \left[\begin{array}{c}
\bm{K}_1 \\
\bm{K}_2
\end{array}\right] \bm{u}\approx\left[\begin{array}{c}
m\left(\ddot{\bm{p}}_c+\bm{g}\right) \\ 
\bm{I}_G\dot{\bm{\omega}}
\end{array}\right]
    \end{aligned} $}
\end{equation}
with
\begin{equation}
\resizebox{0.65\hsize}{!}{$
    \begin{aligned}
        \bm{K}_1 &= \left[\begin{array}{cccccc} \bm{T}_{h} & \bm{T}_{f} & \bm{T}_{f} & \bm{0}_{f} & \bm{0}_{f} & \bm{0}_{f} \end{array}\right] \\
        \bm{K}_2 &= \left[\begin{array}{cccccc} {\bm{d}^{\times}_{ch}} & {\bm{d}^{\times}_{cl}}  & {\bm{d}^{\times}_{cr}} & \bm{L}_h & \bm{L}_f & \bm{L}_f \end{array}\right] \\
    \end{aligned} $}
\end{equation}
We make the approximation of $\frac{d}{dt}(\bm{I}_{G}{\bm{\omega}})\approx \bm{I}_G\dot{\bm{\omega}}$ as the gyroscopic torque term $\bm{\omega}^{\times}(\bm{I}_G\bm{\omega})$ is negligible for small $\bm{\omega}$ and does not affect the dynamics significantly \cite{di2018dynamic}.
$\bm{I}_G$ is computed with \cite{featherstone2014rigid}.
$\bm{T}_h$, $\bm{T}_f$, $\bm{L}_h$ and $\bm{L}_f$ are input selection matrices following frame conventions of $x,y,z$ and $\theta_x, \theta_y, \theta_z$, with
\begin{equation}
    \begin{aligned}
        \bm{T}_h &= 
        \begin{cases}
            \diag\left(\begin{bmatrix} 1 & 1 & 0 \end{bmatrix}\right), & \text{if in contact} \\
            \diag\left(\begin{bmatrix} 0 & 0 & 0 \end{bmatrix}\right), & \text{otherwise}
        \end{cases}, \\
        \bm{L}_h &=
        \begin{cases}
            \diag\left(\begin{bmatrix} 1 & 0 & 0 \end{bmatrix}\right), & \text{if in contact} \\
            \diag\left(\begin{bmatrix} 0 & 0 & 0 \end{bmatrix}\right), & \text{otherwise}
        \end{cases}, \\
        \bm{T}_f &= \diag\left(\begin{bmatrix} 1 & 1 & 1 \end{bmatrix}\right), \\
        \bm{L}_f &= \diag\left(\begin{bmatrix} 0 & 1 & 1 \end{bmatrix}\right).
    \end{aligned}
\end{equation}
We rationalize the choice of hand inputs with the intuition that when pushed, a robot would use its hands to generate a force in the transverse plane and momentum to control its roll rotation.
It is noted that the foot input selection matrix is for the stance foot.
The relationship between the world angular velocity $\bm{\omega}$ and the change rate of the body orientation $\bm{\theta}=[\theta_x,\theta_y,\theta_z]$ can be defined as in \cite{di2018dynamic} with small angle assumptions on the roll and pitch angles:
\begin{equation}
    \frac{d}{dt}\bm{\theta}= \bm{R}_{z}^{-1}\bm{\omega}
\end{equation}
where 
\begin{equation}
\resizebox{0.55\hsize}{!}{$
    \bm{R}_{z}^{-1}=\left[\begin{array}{ccc}
\cos (\theta_z) & \sin (\theta_z) & 0 \\
-\sin (\theta_z) & \cos (\theta_z) & 0 \\
0 & 0 & 1
\end{array}\right]
$}
\end{equation}

Combining the translation and orientation dynamics, we write the ROM as
\begin{equation}
\resizebox{0.6\hsize}{!}{$
\frac{d}{d t}\left[\begin{array}{c}
\bm{\theta} \\
\bm{p}_c \\
\bm{\omega} \\
\dot{\bm{p}}_c
\end{array}\right]=\bm{A}_{{c}}\left[\begin{array}{c}
\bm{\theta} \\
\bm{p}_c \\
\bm{\omega} \\
\dot{\bm{p}}_c
\end{array}\right]+\bm{B}_{{c}} \bm{u}-\left[\begin{array}{c}
\bm{0} \\
\bm{0} \\
\bm{0} \\
\bm{g}
\end{array}\right]$}
\end{equation}
where
\begin{equation}
\resizebox{0.5\hsize}{!}{$   \bm{A}_{{c}}=\left[\begin{matrix}
\bm{0}_{3 \times 3} & \bm{0}_{3 \times 3} & \bm{R}_z^{-1} & \bm{0}_{3 \times 3} \\
\bm{0}_{3 \times 3} & \bm{0}_{3 \times 3} & \bm{0}_{3 \times 3} & \bm{I}_{3 \times 3} \\
\bm{0}_{3 \times 3} & \bm{0}_{3 \times 3} & \bm{0}_{3 \times 3} & \bm{0}_{3 \times 3} \\
\bm{0}_{3 \times 3} & \bm{0}_{3 \times 3} & \bm{0}_{3 \times 3} & \bm{0}_{3 \times 3}
\end{matrix}\right]$}
\end{equation} and
\begin{equation}
 \resizebox{0.95\hsize}{!}{$   \bm{B}_{{c}}=\left[\begin{matrix}
\bm{0_{3 \times 3}} & \bm{0_{3 \times 3}} & \bm{0_{3 \times 3}} & \bm{0_{3 \times 3}} & \bm{0_{3 \times 3}} & \bm{0_{3 \times 3}}   \\
\bm{0_{3 \times 3}} & \bm{0_{3 \times 3}} & \bm{0_{3 \times 3}} &  \bm{0_{3 \times 3}} & \bm{0_{3 \times 3}} & \bm{0_{3 \times 3}} \\
\bm{I}_G^{-1}{\bm{d}^{\times}_{ch}} & \bm{I}_G^{-1} {\bm{d}^{\times}_{cl}} & \bm{I}_G^{-1}{\bm{d}^{\times}_{cr}} & \bm{I}_G^{-1} \bm{L}_h & \bm{I}_G^{-1} \bm{L}_f & \bm{I}_G^{-1} \bm{L}_f   \\
m^{-1} \bm{T}_h & m^{-1} \bm{T}_f & m^{-1} \bm{T}_f &  \bm{0_{3 \times 3}} & \bm{0_{3 \times 3}} & \bm{0_{3 \times 3}} \\
\end{matrix}
\right] $}
\rule[-20pt]{0pt}{20pt}
\end{equation}
To better formulate the dynamics for MPC optimization, we incorporate gravity $\bm{g}$ into the state $\bm{x}=[\bm{\theta}, \bm{p}_c, \bm{\omega}, \dot{\bm{p}}_c, -\bm{g}]$ and rewrite the ROM as 
\begin{equation}
\label{eq:srb_continuous_dynamics}
\dot{\bm{x}}(t)={\bm{A}_{{ce}}}(\theta_z) \bm{x}(t)+{\bm{B}}_{{ce}}\left(\bm{d}_{ch},\bm{d}_{cl},\bm{d}_{cr}\right) \bm{u}(t)
\end{equation}
where ${\bm{A}}_{{ce}}$ and ${\bm{B}}_{{ce}}$ are extended with the gravity term:
\begin{equation}
    \bm{A}_{ce}=\left[ \bm{A}_c \, \middle| \, \begin{array}{c}
     \bm{0_{3 \times 3}} \\  \bm{0_{3 \times 3}} \\  \bm{0_{3 \times 3}} \\  \bm{I_{3 \times 3}}
    \end{array} \right],
    \bm{B}_{ce}=\left[ \bm{B}_c \, \middle| \, \begin{array}{c}
     \bm{0_{3 \times 3}} \\  \bm{0_{3 \times 3}} \\  \bm{0_{3 \times 3}} \\  \bm{0_{3 \times 3}}
    \end{array} \right]
\end{equation}
\begin{figure*}[htb]
    \centering
    \includegraphics[width=\linewidth]{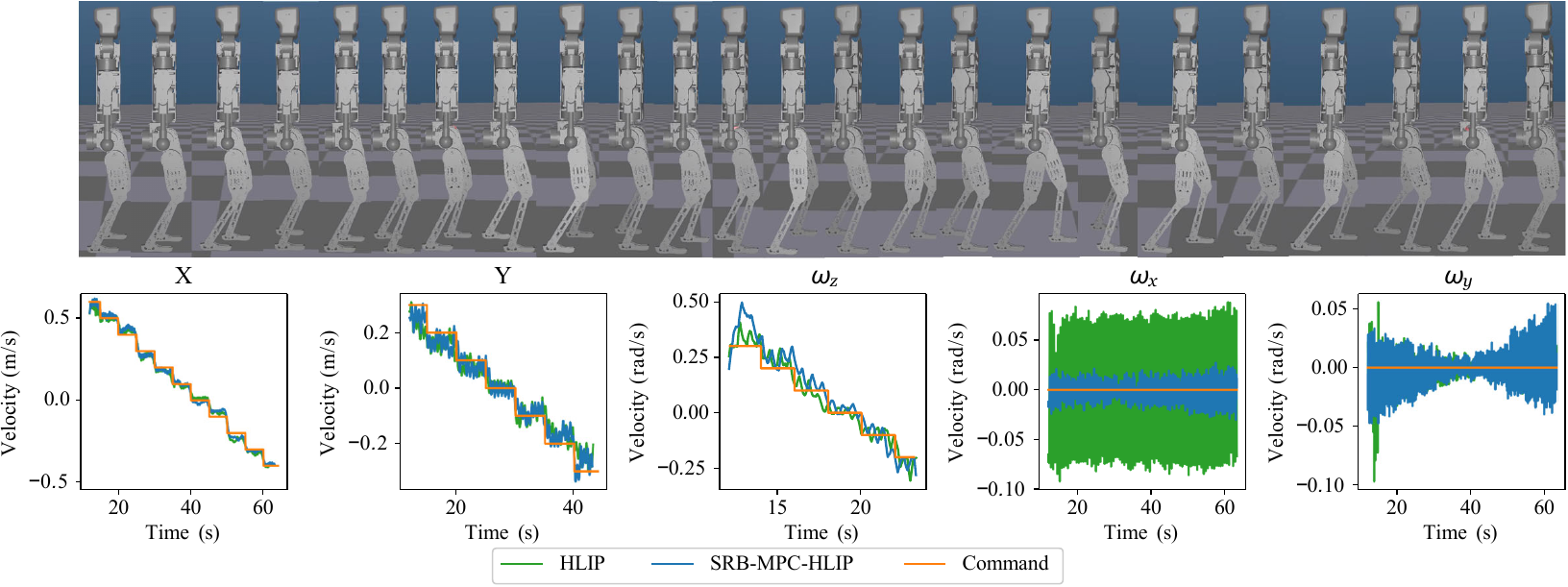}
    \caption{Tracking performance of the pure HLIP controller and the SRB-MPC-HLIP controller. The proposed controller can maintain adequate command tracking and better stabilization of the torso orientation. The snapshot illustrates the robot accelerating to a sagittal velocity of $0.5$ m/s.
    }
    \label{fig:tracking}
\end{figure*}
\subsection{Model Predictive Control}
By discretizing the continuous-time dynamics from \eqref{eq:srb_continuous_dynamics}, we can represent the linearized dynamics in the form of 
\begin{equation}
\bm{x}[i+1]={\bm{A}_d}[i] \bm{x}[i]+{\bm{B}_d}[i] \bm{u}[i]
\end{equation}
with ${\bm{A}_d}$ and ${\bm{B}_d}$ being the discretized zero-hold form of $\bm{A}_{{ce}}$ and $\bm{B}_{{ce}}$ at each time step. 
It is noted that if there is a large deviation from the reference trajectory, the model will become inaccurate, but as discussed in \cite{di2018dynamic}, ${\bm{B}_d}[i]$ is calculated from the current robot states and is always correct. 
As the SRB-MPC runs at a high-enough frequency ($\geq 250$ hz), it will recompute the reference trajectory to account for disturbances. 
We formulate the SRB-MPC as in \ref{ssec:MPC-II}.
It solves for optimal ground reaction forces and moments while satisfying the dynamic constraints and the following contact constraints:
\begin{equation} \label{eq:constraints}
\resizebox{0.95\hsize}{!}{$
\begin{aligned}
    -\mu \bm{F}_{(l,r)z} \leq \bm{F}_{(l,r)x} \leq \mu \bm{F}_{(l,r)z} &\hspace{4pt} \text{(foot sagittal friction cone)}\\
    -\mu \bm{F}_{(l,r)z} \leq \bm{F}_{(l,r)y} \leq \mu \bm{F}_{(l,r)z} &\hspace{4pt} \text{(foot frontal friction cone)}\\
    0 \leq \bm{F}_{(l,r)z} \leq \bm{F}_{f_{\text{max}}} &\hspace{4pt} \text{(max foot supporting force)}\\
    -\mu \bm{F}_{hy} \leq \bm{F}_{hx} \leq \mu \bm{F}_{hy} &\hspace{4pt} \text{(hand sagittal friction cone)}\\
    -\mu \bm{F}_{hy} \leq \bm{F}_{hz} \leq \mu \bm{F}_{hy} &\hspace{4pt} \text{(hand traversal friction cone)}\\
    \bm{F}_{h_{\text{min}}} \leq \bm{F}_{hy} \leq \bm{F}_{h_{\text{max}}}  &\hspace{4pt} \text{(hand supporting force)}\\
\end{aligned}
$}
\end{equation}
where $\bm{F}_{f_{\text{max}}}$, $\bm{F}_{h_{\text{min}}}$ and $\bm{F}_{h_{\text{max}}}$ are dynamically adjusted based on the current contact states and whether to activate the recovery controller inferred from forward kinematics and SRB-MPC projections as discussed in \ref{ssec:recovery}.
\subsection{QP Formulation}
The SRB-MPC problem can be reformulated as a QP as follows:
\begin{equation}
\resizebox{0.5\hsize}{!}{$
\begin{aligned}
\min_{\bm{U}} & \frac{1}{2}\bm{U}^T\bm{H}\bm{U} + \bm{U}^T\bm{m} \\
\text{subject to} & \quad \bm{c}_{\text{min}} \leq \bm{C}\bm{U} \leq \bm{c}_{\text{max}} \\
\end{aligned}$}
\end{equation} 
with $\bm{U} = [\bm{u}_0^T, \bm{u}_1^T, \cdots, \bm{u}_{N-1}^T]^T$ being the stacked decision variables, $\bm{C}$ being the constraint matrix for (\ref{eq:constraints}), and 
\begin{equation}
\begin{aligned}
\bm{H} &= 2\left(\bm{B}_{{qp}}^{\top} \bm{Q} \bm{B}_{{qp}}+\bm{R}\right)
\\
\bm{m} &= 2 \bm{B}_{{qp}}^{\top} \bm{Q}\left(\bm{A}_{{qp}} \bm{x}_0-\bm{x}_{\text{ref}}\right) \\
\end{aligned}
\end{equation}
where $\bm{Q}$ is the weighting matrix for state deviations, $\bm{R}$ is the regularization matrix for input magnitude, $\bm{A}_{{qp}}$ and $\bm{B}_{{qp}}$ are the stacked state transition matrices, and $\bm{x}_{\text{ref}}$ is the reference trajectory.
We solve the QP with qpOASES\cite{ferreau2014qpoases}.
\subsection{Stance leg trajectory}
The resulting optimal solution $\bm{U}^*$ is then used to compute the optimal force and moment that the feet need to exert on the ground and is mapped to joint torques with respective Jacobians as calculated with \cite{featherstone2014rigid}.

\begin{equation}
    \bm{\tau}_{d_{(h,l,r)}} = \bm{J}^T_{(h,l,r)}\bm{u}_{(h,l,r)}
\end{equation}
where
\begin{equation}
    \bm{J}_{(h,l,r)} = \left[\begin{array}{ccc}\bm{J}_{(h,l,r)}^p\bm{T}_{(h,l,r)} & \bm{J}_{(h,l,r)}^o\bm{L}_{(h,l,r)}\end{array}\right]
\end{equation}
where $\bm{J}_{(h,l,r)}^p$ and $\bm{J}_{(h,l,r)}^o$ are the positional and orientation Jacobians, and $\bm{T}_{(h,l,r)}$, $\bm{L}_{(h,l,r)}$ are the selection matrices for the hand, left foot, and right foot defined in \ref{sub:dynamics}.
It is noted that during normal locomotion, we enforce no force on the hand, thus $\mathbf{u}_h$ would be $\mathbf{0}$.
\begin{figure*}[htbp]
    \centering
    \includegraphics[width=\textwidth]{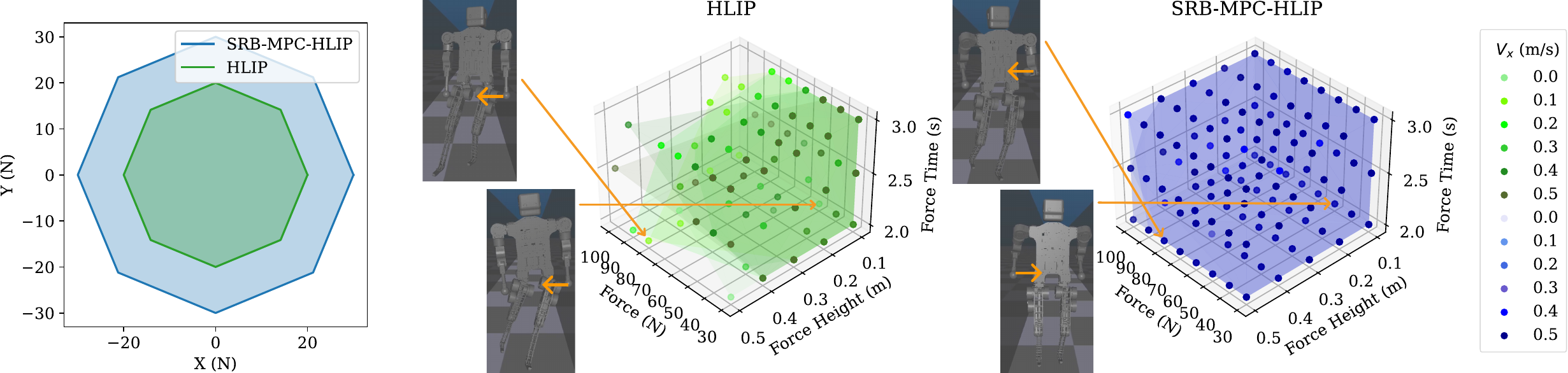}
    \parbox[t]{\textwidth}{%
      \hspace{2.1cm}  (a) \hspace{5.3cm}
         (b) \hspace{5.4cm}
        (c)  
    }
    \caption{(a) Worst-case all-direction safeset of both controllers when pushed while stepping in place with no environmental assistance. The proposed controller has a $225\%$ larger safeset (b) Safeset of the HLIP controller covering $16.8\%$ of total points tested. (c) Safeset of the SRB-MPC-HLIP controller covering $70.6\%$ of total points tested, a $420\%$ increase over the HLIP controller. Snapshots of one easy scenario ($40$N force at $0.1$m above the CoM) and one hard scenario ($80$N force at $0.5$m above the CoM) are also illustrated. It can be seen that the proposed SRB-MPC-HLIP controller maintains better torso stability. The orange arrows in the snapshots show the hand contacts.}
    \label{fig:safesets}
\end{figure*}

\begin{figure}[htbp]
  \centering
   \centering
    \includegraphics[width=\linewidth]{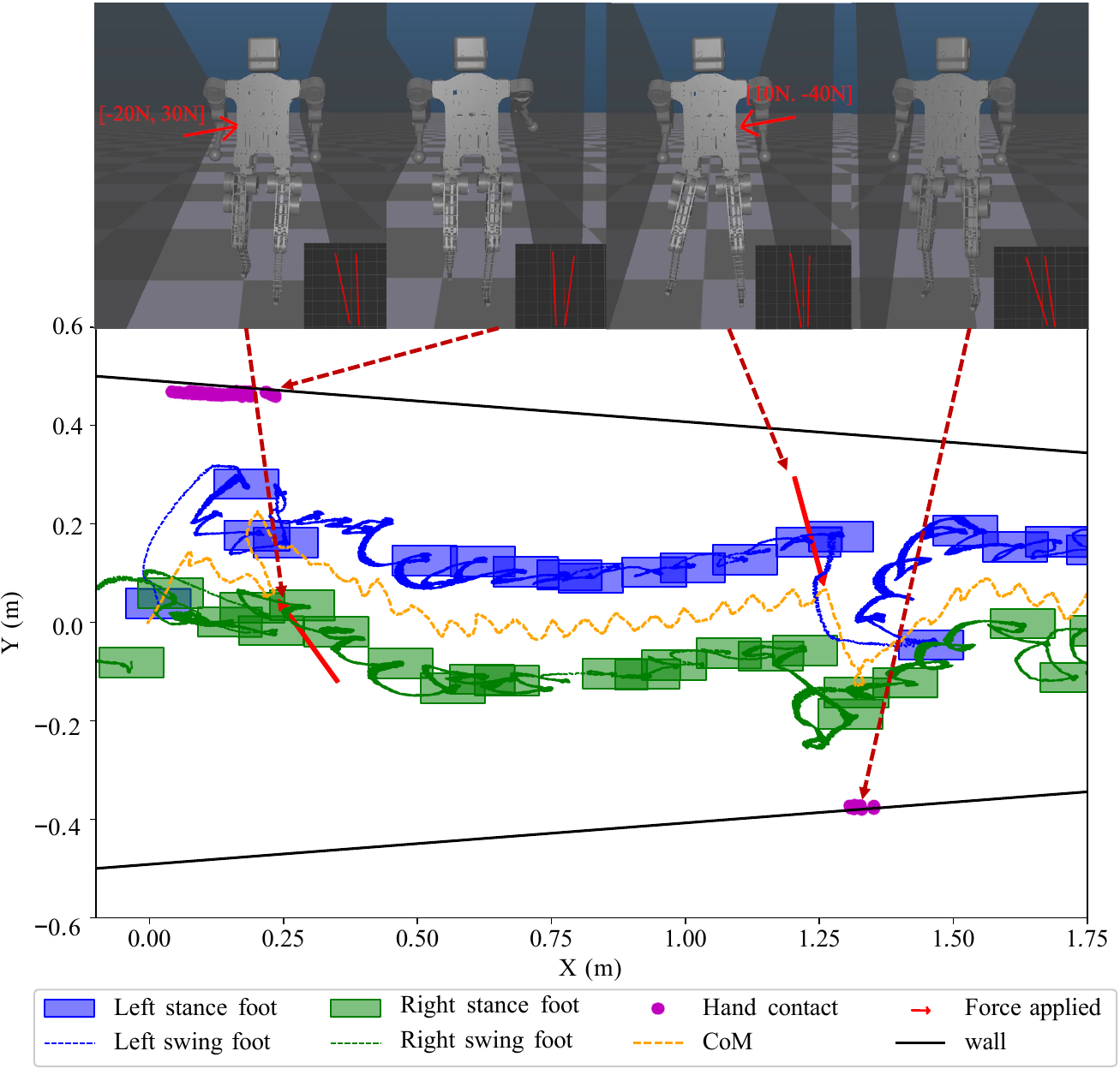}
    \caption{Snapshots from simulated experiments,, the robot is pushed with $[-20\text{N},30\text{N}]$ from the right at $2s$ and $[10\text{N}, -40\text{N}]$ from the left at $12$s. The wall is slanted inwards at 5 degrees. The robot was successful in utilizing its arms to brace against the walls to recover stable walking.}
    \label{fig:robust}
  \label{fig:mainfigure}
\end{figure}

\subsection{Swing Leg Trajectory} 
As we apply the zero constraint to the swing leg, it does not require any ground reaction force. Thus, the swing leg trajectory is generated based on \cite{xiong20223}.

Subject to a constant step time, $T_{\text{SSP}}$, we formulate the linear S2S dynamics as the states at the start and termination of each step: 

\begin{equation}
 \resizebox{0.95\hsize}{!}{$   
       (x,y)_{k+1}^{\text{HLIP}} = \underbrace{e^{A_{\text{SSP}}T_{\text{SSP}}}}_A(x,y)_{k}^{\text{HLIP}} + \underbrace{e^{A_{\text{SSP}}T_{\text{SSP}}}\begin{bmatrix}
        -1 \\ 0
    \end{bmatrix}}_B u_k^{\text{HLIP}} 
    $}
\end{equation}

In order to find $u_k$, the nominal step length, we use the characterized periodic orbits as described in \cite{xiong20223}.

Finally, HLIP closes the loop on the step length using the following control law:
\begin{equation}
    u_k = u_k^{\text{HLIP}} + K((x,y)^{\text{robot}}_{k, \text{SSP}}-(x,y)_k^{\text{HLIP}})
\end{equation}
where $K$ is a gain matrix stabilizing $A-BK$. 

We fit a Bezier polynomial, $b(t)$, to define the foot trajectory such that the foot impacts the ground at the requisite step length and achieves a desired step height $z_{\mathrm{sw}}^{\max }$.

\begin{equation}
    \label{eq:swf_x/y}
    (x,y)_{\mathrm{sw}}(t)=\left(1-b(t)\right) (x,y)_{\mathrm{sw}}^{+}+b(t) u_{x}
\end{equation}

\begin{equation}
    \label{eq:swf_z}
    z_{\mathrm{sw}}(t)=b\left(t, z_{\mathrm{sw}}^{\max }, z_{\mathrm{sw}}^{\mathrm{min}}\right)
\end{equation}  
Similarly as in \cite{ghansah2024dynamic}, instead of using \eqref{eq:swf_x/y} directly, we replace the post-impact state with the current swing foot state to smooth out the swing foot trajectories to avoid excessive oscillations.
With the desired swing leg trajectory, we use inverse kinematics to generate the desired joint trajectories $[\bm{P}_d, \bm{V}_d]$ for the swing leg and track them with a PD controller.
As stated, we also increase the stepping frequency when a push is detected to better facilitate adjustments of the support polygon.

\subsection{Push Detection \& Arm Trajectory}
With the aforementioned MPC setup, we perform push detection by looking at the deviation of the robot states from the nominal ones when we propagate the states forward.
This allows us to tune the conditions based on different robots.
After a push is detected, we calculate the distance to the nearest supporting surface and use differential inverse kinematics to move the arm so that the hand makes contact with the surface. 
When contact is made, we determine the position of contact in the robot frame with forward kinematics.
Then we update the MPC to include optimizing for the hand contact force.
After we have the optimal $\mathbf{u}_h$, we calculate the feed-foward torques by utilizing the arm jacobian, and apply them to the relevant arm joints.

\section{Experiments and Evaluations}\label{sec:experiment}

\subsection{Hardware \& Simulation Specifications}
We use a humanoid robot with CoM height 0.7 meters, and weighing 20kg with 24 DoFs, consisting of 2 5-DoF legs denoted as $\bm{q}_\text{F}={q^{\text{L/R}}_{\text{F}_{1,2,3,4,5}}}$ with each leg having hip yaw, hip roll, hip pitch, knee pitch, and ankle pitch motors in the order as listed, and 2 4-DoF arms denoted as $\bm{q}_\text{H}={q^{\text{L/R}}_{\text{H}_{1,2,3,4}}}$ with each arm having shoulder pitch, shoulder roll, shoulder yaw, and elbow pitch motors, and a floating base with 6 DoFs consisting of the CoM states $\bm{q}_{{\text{CoM}}} = q_{x,y,z,\theta_x,\theta_y,\theta_z}$.  
We benchmark the performance of the proposed controller in the high-fidelity simulation with the MuJoCo physics engine \cite{todorov2012mujoco} with default friction settings.
We also employ the use of a sensor simulation \cite{vaughan2008massively} with matching environments to simulate the LiDAR required for the framework.
\subsection{Tracking Performance}
In order to show that there is no degradation from our SRB-MPC-HLIP controller as opposed to only using a pure HLIP-based controller, we perform separate stepped command tests in the x, y, and yaw velocities.
The performance of both controllers is shown in Fig.\ref{fig:tracking}.
It can be observed that the SRB-MPC-HLIP controller retains adequate tracking performance while being more capable at maintaining roll tracking, with it limited to $[-0.025,0.025]$ rad/s as opposed to the $[-0.075,0.075]$ rad/s of the HLIP controller.
This is due to the MPC optimizing for zeroing angular velocities of the torso.
We also perform perturbation checks while stepping in place and find that in the worst case, the pure HLIP controller can only reject up to 20N of force applied to its CoM while the SRB-MPC-HLIP controller can reject up to 30N in all directions, as shown in Fig. \ref{fig:safesets}(a).

\subsection{Push Recovery Performance}
To evaluate the effectiveness of our unified controller in rejecting large perturbations by utilizing its surrounding environment, we perform exhaustive simulated push tests over a range of $[30,100]$N, with a step size of $10$N. 
The force is applied on both sides of the robot, at a varying height relative to the CoM between $0.1$m to $0.5$m with a step size of $0.1$m as an impulse for $0.2$s to simulate a perturbation force on the torso. 
The robot is directed to walk for $2$s at a command x velocity between $[0.0, 0.5]$ m/s, with a step size of $0.1$ m/s, before the push is applied between $2$s and $3$s with a step size of $0.5$s to account for all combinations of stance and swing feet.
We categorize failure as its torso making contact with the ground or the surrounding walls, its torso roll or pitch angle exceeding $0.3$ rad in magnitude, its CoM dropping for more than $0.1$m below its nominal height, or its shoulders coming into prolonged contact with the walls.
In total, 1440 data points were collected and the nominal HLIP controller can only handle $16.8\%$ of the cases while the SRB-MPC-HLIP controller can handle $70.6\%$, providing a $420\%$ safeset size increase.
The resulting safe set is illustrated in Fig. \ref{fig:safesets}(b) and Fig. \ref{fig:safesets}(c).
As seen, the proposed method can recover better from high-speed and high-force pushes and is robust to the robot configuration at the time of the push.
To further illustrate the effectiveness of the proposed method and  show robustness to different perturbations and variations in the environment, we also test the controller in scenarios such as with angled walls and different force directions in Fig. \ref{fig:robust}, with the first push being $[-20\text{N},30\text{N}]$ on the right, the second push being $[10\text{N}, -40\text{N}]$ on the left, and the walls are slanted inwards at 5 degrees.
As seen, the robot can successfully recover from the perturbation forces and continue walking, further proving the viability of our controller.

\section{Conclusion}\label{sec:concusion}
We have demonstrated a unified model predictive control framework capable of robust locomotion and push recovery.
It employs the Single Rigid Body model and the Hybrid Linear Inverted Pendulum model to achieve robust perturbation rejection by interacting with the environment and modulating its step period and next footstep location. 
The proposed SRB-MPC-HLIP controller was validated extensively in a high-fidelity simulation on a humanoid robot, demonstrating significant improvement in terms of perturbation force rejection and better tracking performance.
As future work, we envision using a learned classifier to detect and predict pushes, deploying the framework on hardware, extending it to include internal disturbances caused by the movements of the limbs, and adapting to varying kinds of environmental surfaces and objects. We also hope to deploy the method on different humanoids to exhibit its generalizability.

\balance
\bibliographystyle{IEEEtran}
\bibliography{references}

\end{document}